\pdfoutput=1

\documentclass[11pt]{article}

\usepackage[table]{xcolor}
\usepackage{EMNLP2022}

\usepackage{times}
\usepackage{latexsym}
\usepackage{amssymb}
\usepackage{booktabs}
\usepackage{graphicx}
\usepackage{subcaption}

\usepackage[T1]{fontenc}

\usepackage[utf8]{inputenc}

\usepackage{microtype}

\usepackage{inconsolata}

\title{Enhancing Crisis-Related Tweet Classification with Entity-Masked Language Modeling and Multi-Task Learning}

\author{Philipp Seeberger \and Korbinian Riedhammer \\
  Technische Hochschule Nürnberg Georg Simon Ohm \\
  \texttt{\{philipp.seeberger,korbinian.riedhammer\}@th-nuernberg.de}}

\begin{document}
\maketitle
\begin{abstract}
Social media has become an important information source for crisis management and provides quick access to ongoing developments and critical information.
However, classification models suffer from event-related biases and highly imbalanced label distributions which still poses a challenging task.
To address these challenges, we propose a combination of entity-masked language modeling and hierarchical multi-label classification as a multi-task learning problem.
We evaluate our method on tweets from the TREC-IS dataset and show an absolute performance gain w.r.t. F1-score of up to 10\% for actionable information types.
Moreover, we found that entity-masking reduces the effect of overfitting to in-domain events and enables improvements in cross-event generalization.
Our source code is publicly available on GitHub.\footnote{\url{https://github.com/th-nuernberg/crisis-tapt-hmc}}
\end{abstract}

\section{Introduction}

Messages on social media during disaster events have become an important information source in crisis management \citep{reuter_2018}.
In contrast to traditional sources (e.g., official news), social media posts immediately provide details about developments, first-party observations, and affected people in an ongoing emergency situation \citep{sakaki_2010}.
Having access to this information is crucial for developing situational awareness and supporting relief providers, government agencies, and other official institutions \citep{kruspe_2021}.

One key challenge poses the information refinement of high-volume social media streams which requires automatic methods for reliable detection of relevant content \citep{kaufhold_2021}.
Most recent work has focused on binary, multi-class, and multi-label text classification techniques to classify posts into coarse (e.g., \textit{Relevant}, \textit{Irrelevant}) or fine-grained (e.g., \textit{InfrastructureDamage}, \textit{MissingPeople}) categories composed of flattened or hierarchical structures \citep{crisismmd_2018,alam_2021,trec_is_2021}.

\begin{figure}[t]
    \includegraphics[width=\linewidth]{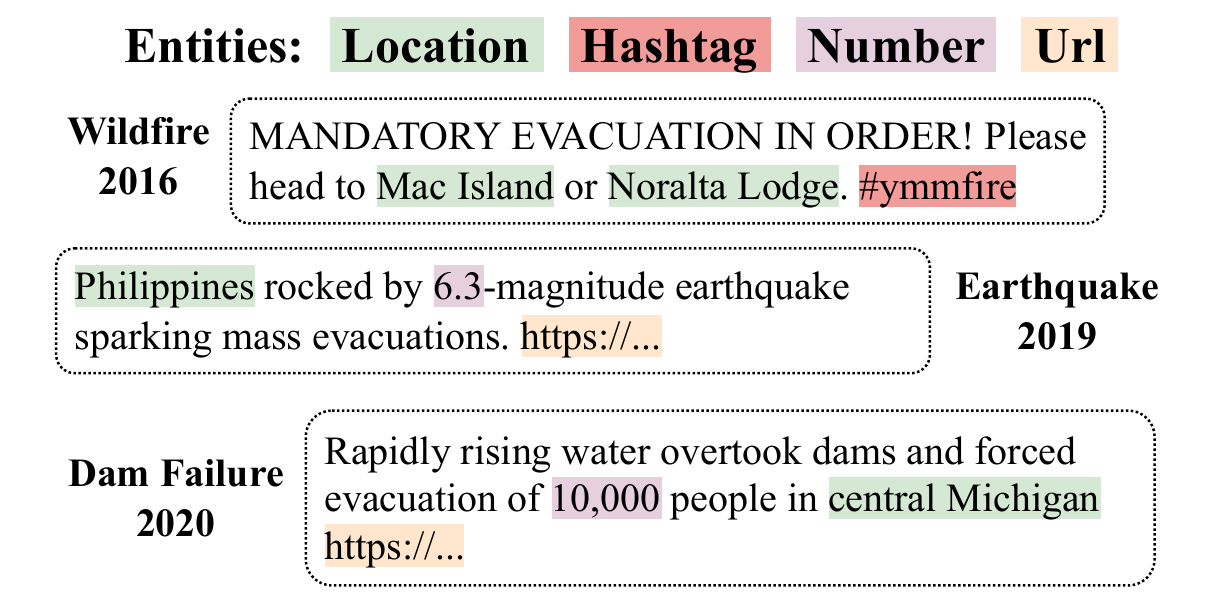}
    \caption{Example tweets of several disasters over time, annotated with entitites. The short posts are mostly biased towards specific events.}
    \label{fig:intro:tweets}
\end{figure}

Another challenge in Natural Language Processing (NLP) is the nature of data prevalent in social media and microblogging platforms.
For example, most works in the crisis-related domain focus on Twitter data \citep{kruspe_2021} which inherits properties such as short texts (280 characters limitation per tweet), less contextual information, hashtags, and noise (e.g., misspellings, emojis) \citep{wiegmann_2020,zahera_2021}.
According to \citet{sarmiento_2021}, different types of disasters (e.g., flood, wildfire) can be identified by only a few text-based features.
However, event-related biases and entities as shown in \figurename~\ref{fig:intro:tweets} prevent models from generalizing to unseen disaster events and therefore degrade w.r.t. detection performance.

To circumvent this problem, approaches such as adversarial training \citep{medina_2020}, domain adaptation \citep{alam_2018}, and hierarchical label embeddings \citep{miyazaki_2019} have been proposed but suffer from mixed event types, assume unlabeled data or require semantic label descriptions.
Contrary to this work, we aim to enhance the detection of rare actionable information for unseen events by masking out entities, applying adaptive pre-training, and incorporating the hierarchical structure of labels. 

\paragraph{Contributions} Our main contributions are as follows:
(1) We introduce an adaptive pre-training strategy based on entity-masking.
(2) We incorporate the hierarchical structure of labels as multi-task learning (MTL) problem.
(3) We empirically show that our approach improves generalization to new events and increases detection performance for actionable information types.

\section{Related Work}

\paragraph{Crisis Tweet Classification}

Besides conventional detection approaches such as filtering \citep{kumar_2011} or crowdsourcing \citep{poblet_2014}, machine learning has received much attention in this area.
Researchers experimented with several methods such as Naive Bayes, Support Vector Machines, and Decision Trees either with term-frequency features \citep{habdank_2017} or static embeddings \citep{kejriwal_2019}.
More recently, the combination of Word2Vec \cite{mikolov_2013} with Convolutional and Recurrent Neural Networks achieved remarkable improvement in this field \citep{kersten_2019,snyder_2019}.
Due to the success of Transformers \citep{vaswani_2017} and the follow-up language models \cite{devlin_2019}, most works have been built upon this and outperformed previous approaches \citep{alam_2021,wang_2021a}.

\begin{table}[t]
    \centering
    \begin{tabular}{lcc}
        \hline
         & \textbf{Train} & \textbf{Test} \\
        \hline
        Event Ids  & 1 - 52 & 53 - 75 \\
        \# Events & 51  & 21 \\
        \# tweets & 50,412 & 22,003 \\
        \hline
        \textit{Upper classes} \\
        \# Report (14) & 30,389 & 16,059 \\
        \# Other (5) & 32,105 & 10,709 \\
        \# CallToAction (3) & 1,458 & 389 \\
        \# Request (3) & 683 & 144 \\
    \hline
    \end{tabular}
    \caption{Overview of the dataset split; the values within the brackets of the upper classes corresponds to the number of unique low-level information types.}
    \label{table:dataset}
\end{table}

\paragraph{Adaptive Pre-Training}

Transfer learning with language models essentially contributes to state-of-the-art results in a variety of NLP tasks \citep{devlin_2019,roberta_2019,electra_2020}.
Typically, such language models follow the three training steps \citep{howard_2018,ben_david_2020}: 
(1) Pre-training on massive corpora; 
(2) Optional pre-training on task-specific data; 
(3) Supervised fine-tuning on target tasks.
However, the second step is often neglected due to computational constraints whereby adaptive pre-training has shown to be effective \citep{howard_2018}.
Hence, \citet{gururangan_2020} introduced domain-adaptive pre-training (DAPT) and task-adaptive pre-training (TAPT) which cover continual pre-training on corpora tailored for a specific task.
Moreover, strategies such as adding special tokens for tweets \citep{bertweet_2020,wiegmann_2020} or additional masked language modeling (MLM) approaches \citep{ben_david_2020} have been proven beneficial. 

\begin{figure*}[t]
    \begin{subfigure}[h]{0.45\linewidth}
    \includegraphics[height=55mm, width=\linewidth]{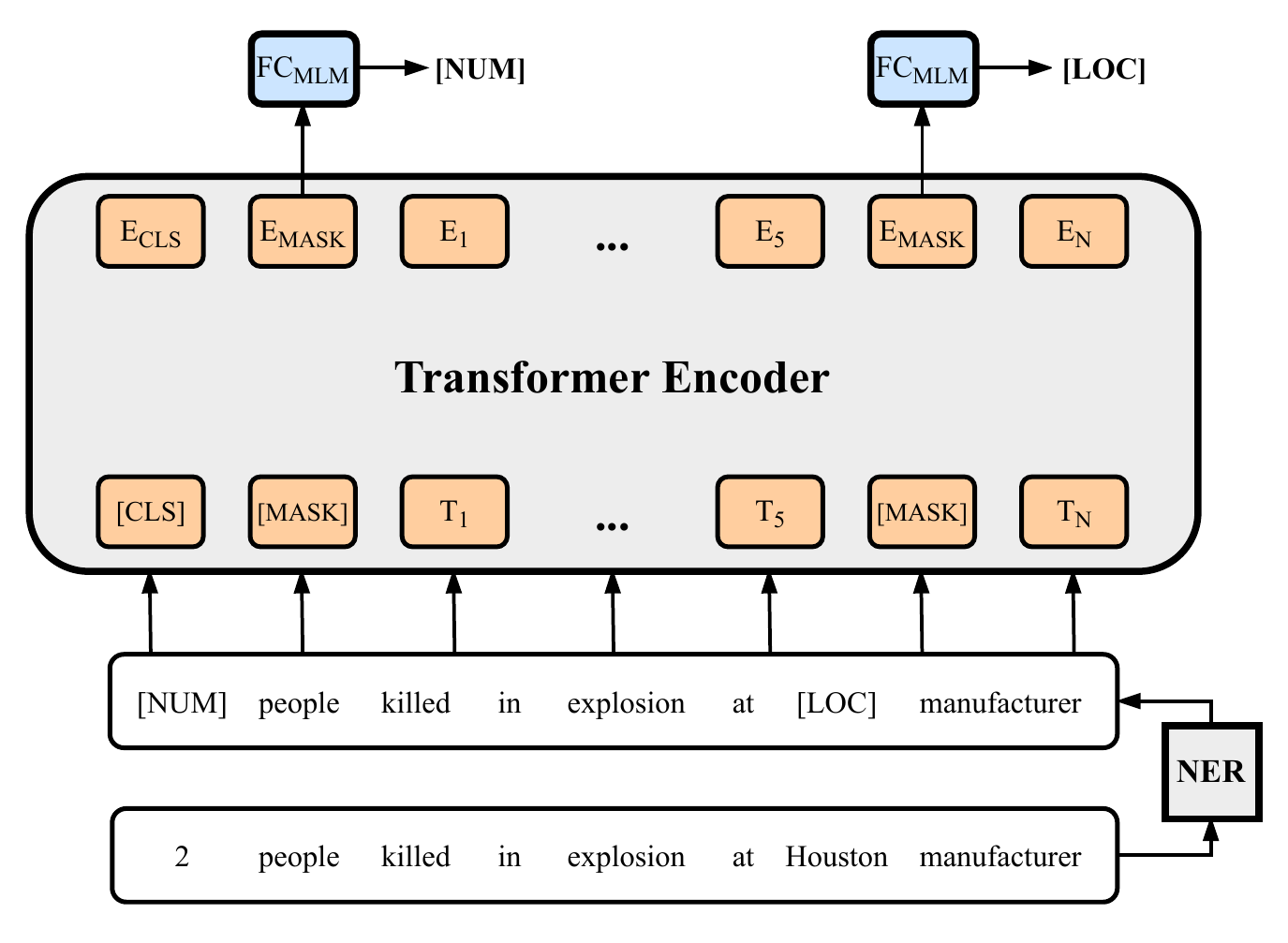}
    \caption{Entity-masked language modeling}
    \label{fig:methods:emlm}
    \end{subfigure}
    \hfill
    \begin{subfigure}[h]{0.54\linewidth}
    \includegraphics[height=55mm, width=\linewidth]{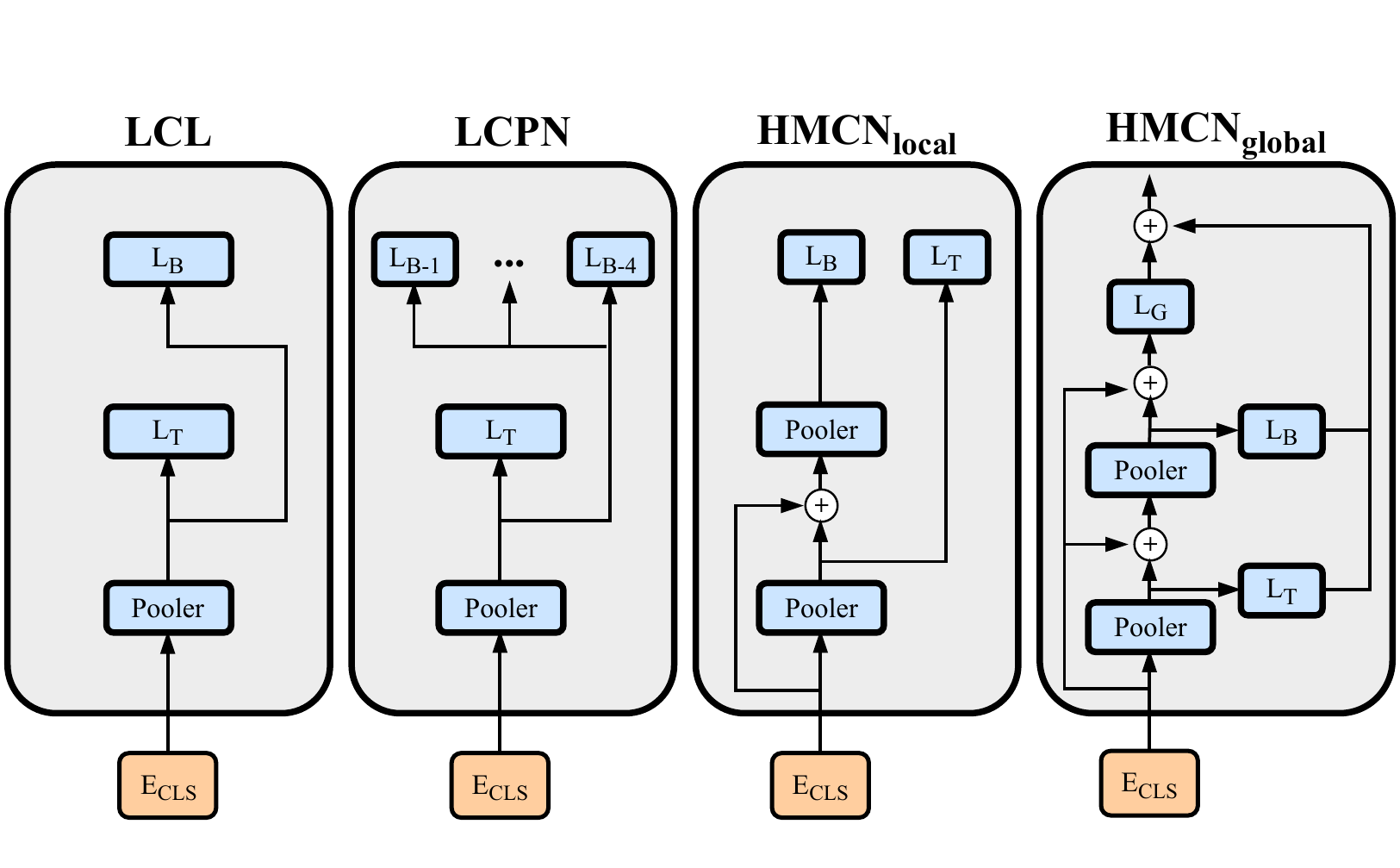}
    \caption{Multi-task classification heads}
    \label{fig:methods:hmc}
    \end{subfigure}%
    \caption{Illustration of the concepts E-MLM with named entity recognition (NER) and MTL. FC$_{MLM}$ represents the prediction head for MLM. The classification heads will be placed on top of the pre-trained encoder. The building blocks Pooler, L$_{T}$, L$_{B}$ and L$_{G}$ are fully connected layers and use the CLS token as sentence embedding.} 
    \label{fig:methods}
\end{figure*}

\paragraph{Hierarchical Multi-Label Classification}

Hierarchical multi-label classification (HMC) covers local and global approaches and the combination of both worlds \citep{wehrmann_2018}.
A popular categorization of local methods is the subdivision into local classifier per parent node (LCPN) \citep{dumais_2000}, local classifier per node (LCN) \citep{banerjee_2019}, and local classifier per level (LCL) \citep{wehrmann_2018}.
Hybrid approaches integrate the global part as a particular constraint such as hierarchical softmax \citep{brinkmann_2021} or combine multiple local and global prediction heads \citep{wehrmann_2018}.
Recent work in information type classification introduced label embeddings which utilize the hierarchical structure \citep{miyazaki_2019}.
Finally, the classification can also be viewed as MTL by combining certain loss functions \citep{yu_2021,wang_2021a}.

\section{TREC-IS}

In this work, we mainly focus on the dataset of the shared-task TREC-IS, which represents a collection of annotated crisis-related tweets \citep{trec_is_2021}.
Each tweet belongs to a disaster event and is annotated with high-level information types which are derived from an ontology composed of hierarchical stages.
However, information type labels are only shipped as a two-level hierarchy with four upper classes L$_{T}$  and 25 lower classes L$_{B}$.
Thus, both hierarchy levels represent a multi-label classification task.
Following the TREC-IS track design, we split the dataset into train and test events which corresponds to the TREC-IS 2020B task. 
This split poses a challenging setup due to the requirement of cross-event generalization \citep{wiegmann_2020}.
\tablename~\ref{table:dataset} gives an overview of each split; obviously, the information type distribution is highly imbalanced.
For example, information types with low criticality such as \textit{MultimediaShare} (31.7\%) and \textit{News} (25.4\%) are prevalent.
In contrast, the highly critical information types \textit{MovePeople} (0.9\%) and \textit{SearchAndRescue} (0.4\%) occur only rarely \citep{trec_is_2019}.\footnote{We provide an overview of the labels with some example posts in Appendix \ref{appendix:dataset_details}.}

\section{Method}

As depicted in \figurename~\ref{fig:methods} our approach combines the two concepts entity-masked language modeling (E-MLM) and MTL.
In the following, we briefly describe our method as a combination of those two.

\subsection{Entity-Masked Language Modeling}

Based on adaptive pre-training, we extend on masked language modeling of a transformer encoder pre-trained on a large corpus such as BERT \citep{devlin_2019}.
Here, the mitigation of event-related biases is facilitated by replacing entities -- which are prone to be event-specific  -- with special tokens (see \figurename~\ref{fig:methods:emlm}).
This way we intend to capture disaster-related language patterns independently of the concrete entities.
Following \citet{ben_david_2020}, we further introduce a masking probability $\alpha$ tailored to entities in addition to the standard word masking with probability $\beta$.
That is, with a typically higher probability $\alpha$ we select random entity-tokens such as locations and lower probability $\beta$ random standard subword-tokens. Finally, these selected tokens will be replaced by \textit{[MASK]}, random tokens or the unchanged tokens in order to learn the linguistic patterns related to those entities. 
For the rest of this paper, we rely on the pre-trained BERT$_{BASE}$ as the encoder model and the corresponding default MLM setup for pre-training (\textit{[MASK]} with 80\%, random tokens with 10\%, and unchanged tokens with 10\%).

\subsection{Multi-Task Learning}

The next step represents the fine-tuning of a classification head.
We implement four basic hierarchical multi-label classification approaches as shown in \figurename~\ref{fig:methods:hmc}.
The LCL classification head jointly trains a flattened classification layer for each of the two hierarchy levels.
In contrast, the LCPN model consists of a classification layer for each parent node.
The hierarchical multi-label classification network (HMCN) is adapted from \citet{wehrmann_2018} and introduces a pooling layer on top of the preceding pooling layer.
We experiment with a local and a global variant, whereas the global one additionally consists of a global classification layer.
All pooling and classification layers are composed of a single feed-forward layer with \textit{tanh} and \textit{sigmoid} as activation functions, respectively.
Finally, we minimize the binary cross-entropy $\mathcal{L}_{MTL} = \lambda \mathcal{L}_{L_{T}} + (1 - \lambda) \mathcal{L}_{L_{B}}$ as a weighted loss function whereby $\mathcal{L}_{L_{T}}$ represents the upper classes and $\mathcal{L}_{L_{B}}$ the lower classes loss.

\section{Experiments}

\subsection{Evaluation Metric}

We follow the TREC-IS evaluation scheme: macro-averaged F1-score across information types for the two hierarchy levels in addition to the actionable information types (AIT) \citep{trec_is_2019}.
The latter include rare information types with high priority consisting of: \textit{MovePeople}, \textit{EmergingThreats}, \textit{NewSubEvent}, \textit{ServiceAvailable}, \textit{GoodsServices}, and \textit{SearchAndRescue}.

\subsection{Named Entity Recognition}

As event-specific entities, we use the special tokens \textit{hashtag}, \textit{url}, \textit{person}, \textit{location}, \textit{organization}, \textit{event}, \textit{address}, \textit{phone number}, \textit{date}, and \textit{number}.
All entities except the tokens \textit{hashtag} and \textit{url} are extracted with the Natural Language API of the Google Cloud Platform.\footnote{We extracted the entities on 29 March 2022.}
We manually annotated 300 tweets and calculated a strict F1-score \citep{semeval_2013} of 0.692 which represents a reasonable good result for tweets.

\subsection{Baseline and Hyper-Parameters}

As baseline, we use TF-IDF with Logistic Regression (TF-IDF+LR) and BERT$_{BASE}$ with a single-task classification head.
Furthermore, we apply the standard MLM of BERT in contrast to E-MLM in order to validate the effect of masking entities.
Lastly, we train the MTL model (MTL$_{prio}$) from \citet{wang_2021a} which combines lower classes as classification and priority scores as regression task.
We choose the best hyper-parameters for each model based on a stratified split with a ratio of 90\% for train and 10\% for development data, respectively.
In terms of hyper-parameters, we set $\alpha=0.5$ and $\beta=0.1$ for E-MLM; other parameters were set according to other work, including learning rate of $5e-5$, batch size of \textit{32}, and $\lambda=0.1$ for fine-tuning.
The detailed hyper-parameter selection process is shown in Appendix \ref{appendix:hyperparameters}.

\begin{table}[t]
    \centering
    \begin{tabular}{lccc}
        \hline
        \textbf{Model}  & \textbf{L$_{T}$} & \textbf{L$_{B}$} & \textbf{AIT} \\
        \hline
        \textit{Single-Task} \\
        TF-IDF+LR & 0.657 & 0.499 & 0.462 \\
        BERT$_{BASE}$ & 0.717 & 0.531 & 0.513 \\
        BERT$_{MLM}$ & 0.714 & 0.551 & 0.546 \\
        BERT$_{E-MLM}$ & 0.701 & 0.481 & 0.444 \\
        \hline
    \end{tabular}
    \caption{Overall results on the development set.}
    \label{table:results:development}
\end{table}

\begin{table}[t]
    \centering
    \begin{tabular}{lccc}
        \hline
        \textbf{Model}  & \textbf{L$_{T}$} & \textbf{L$_{B}$} & \textbf{AIT} \\
        \hline
        MTL$^{\ast}_{prio}$ & - & 0.278 & 0.279 \\
        \hline
        \textit{Single-Task} \\
        TF-IDF+LR & 0.460 & 0.201 & 0.168\\
        BERT$_{BASE}$ & \underline{0.553} & 0.269 & 0.236 \\
        BERT$_{MLM}$ & 0.524 & 0.245 & 0.229 \\
        BERT$_{E-MLM}$ & \underline{0.553} & 0.307 & 0.306 \\
        \hline
        \multicolumn{2}{l}{\textit{Multi-Task}} \\
        LCL & 0.548 & \textbf{0.314} & 0.309\\
        LCPN & 0.548 & 0.305 & 0.307 \\
        HMCN$_{global}$ & 0.546 & 0.310 & \underline{0.320}\\
        HMCN$_{local}$ & \textbf{0.558} & \underline{0.312} & \textbf{0.335} \\
        \hline
    \end{tabular}
    \caption{Overall results of information type classification; bold and underlined values indicate the best and second-best results, respectively. $^\ast$We fine-tuned the approach of \citet{wang_2021a} with BERT$_{BASE}$ and without ensembling.}
    \label{table:results}
\end{table}

\begin{figure}[t]
    \center
    \includegraphics[width=\linewidth]{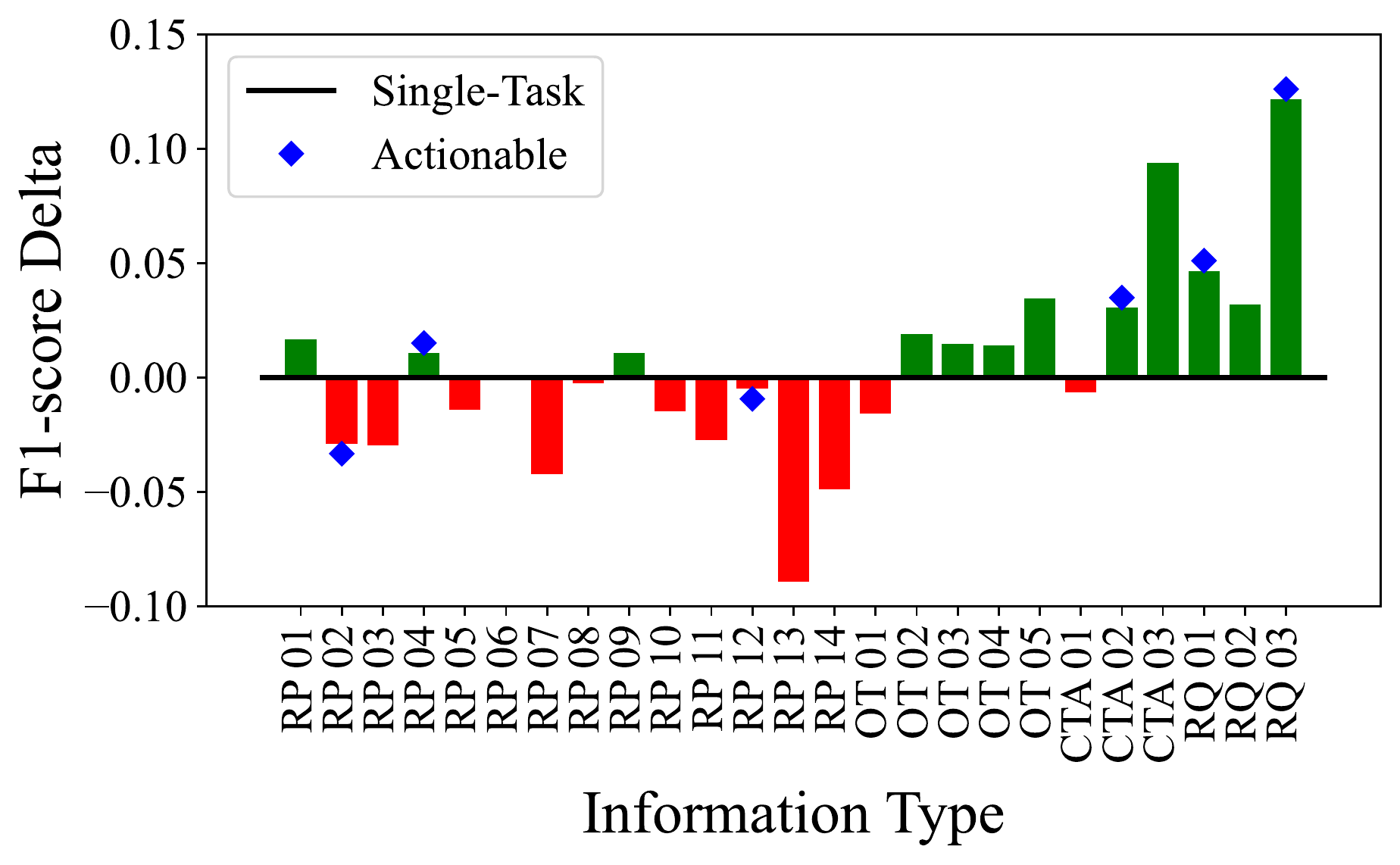}
    \caption{Absolute performance differences w.r.t. F1-score between the single-task and HMCN$_{local}$ model.}
    \label{fig:results:delta}
\end{figure}

\subsection{Results}

In the following, we report the performance for the upper classes L$_{T}$, lower classes L$_{B}$, and AIT.
However, for our evaluation we do not focus on L$_{T}$ since the experiments did not show large differences across all BERT models.
The MTL models are only reported with BERT$_{E-MLM}$.

\paragraph{E-MLM}

\tablename~\ref{table:results} displays the results of all single-task and MTL runs.
For E-MLM, we observe an absolute performance gain w.r.t. F1-score for both L$_{B}$ and AIT by up to 4\% and 7\%, respectively.
To validate the event-generalization effect, we additionally analyzed the development set, as a proxy to estimate the in-domain event performance as shown in \tablename~\ref{table:results:development}.
Contrary to the test set, standard MLM increases the absolute L$_{B}$ performance by 2\% whereas the E-MLM approach drops by 5\% which is a confirmation of our assumption about event-related overfitting.

\paragraph{Multi-Task Learning} 

In terms of MTL, the HMCN$_{local}$ model achieved the best results for AIT. 
Overall the MTL classification outperforms the single-task models for actionable categories and in addition the L$_{B}$ classes except for LCPN. 
We assume that the L$_{T}$ classification objective implicitly clusters the internal representation w.r.t. the high-level information types and therefore mitigates overfitting towards the major classes.
As depicted in \figurename~\ref{fig:results:delta}, the HMCN$_{local}$ model improves the detection of rare actionable information types over the single-task model while at the same time decreasing the performance on the category with the most information types.
This can be caused by the ambiguous label definitions and semantic similarities with other information types \citep{mehrotra_2022}.

\begin{figure}[t]
    \center
    \includegraphics[width=\linewidth]{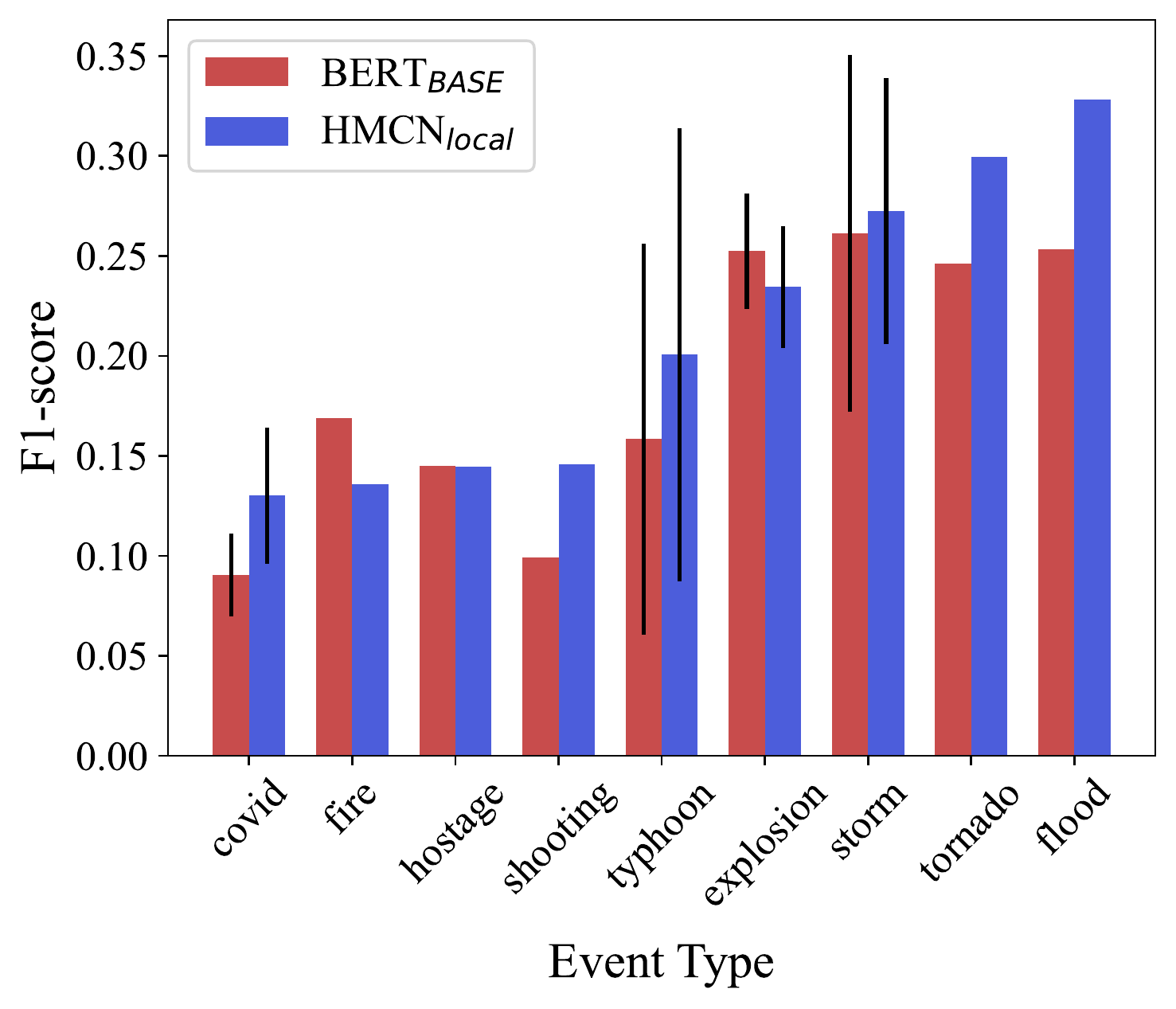}
    \caption{Comparison across event types w.r.t. F1-score between the BERT$_{BASE}$ and HMCN$_{local}$ model. We plot the mean and standard deviation for multiple events within a event type.}
    \label{fig:results:eventtypes}
\end{figure}

\subsection{Analysis of Events} 

In \figurename~\ref{fig:results:eventtypes} we illustrate the model performance for L$_{B}$ across different event types.
For multiple events, we report the mean and standard deviation, respectively.
We observe an increase in performance for the event types \textit{covid}, \textit{shooting}, \textit{typhoon}, \textit{storm}, \textit{tornado}, and \textit{flood} and a small decrease for the event types \textit{fire}, \textit{hostage}, and \textit{explosion}.
As shown by the variance for multiple events, the performance highly differs across specific events.
Surprisingly, the event type \textit{covid} achieved the worst performance for both models despite the existence of three \textit{covid} events within the train data.
These results indicate that even regional differences about the same global event predominantly affect the generalization performance across events.

\subsection{Ablation Study}

As ablation study we removed several proposed components to assess the performance impact of our model.
Thereby, the component entities represents the additional special tokens and replacement within the input text.
As shown in \tablename~\ref{table:results:ablation}, we started with the HMCN$_{local}$ model and demonstrate that entities, MLM and MTL contribute to an increase w.r.t. F1-score for both L$_{B}$ and AIT.
The results indicate that the variant which removes the hierarchical component only degrades the performance for the low-resource actionable information types.
Removing the E-MLM mechanism degrades the model's performance most in our experiments.

\begin{table}[t]
    \centering
    \begin{tabular}{lcccccc}
        \hline
        \textbf{Method}  & \textbf{L$_{T}$} & \textbf{L$_{B}$} & \textbf{AIT} \\
        \hline
        HMCN$_{local}$ & 0.558 & 0.312 & 0.335 \\
        \hline
        - Hierarchy & 0.548 & 0.314 & 0.309 \\
        \hspace{2mm} - Multi-Task & 0.553 & 0.307 & 0.306 \\
        \hspace{4mm} - MLM & 0.529 & 0.276 & 0.242 \\
        \hspace{6mm} - Entities & 0.553 & 0.269 & 0.236 \\
        \hline
    \end{tabular}
    \caption{Overall results of the ablation study.}
    \label{table:results:ablation}
\end{table}

\section{Conclusion and Future Work}

In this work, we identified shortcomings in the field of crisis tweet classification for unseen events.
For the TREC-IS data, we found contrasting effects in terms of pre-training and observed an absolute improvement of up to 3\% w.r.t. F1-score for actionable information types by incorporating the hierarchical structure.
Furthermore, we confirmed the effectiveness of our method based on the shared-task TREC-IS.
Future work includes pre-training on a larger corpus, the mitigation of the trade-off between major and minor classes performances, and to analyse the influence of label semantics.

\section*{Ethical and Societal Implications}

Open Source Intelligence (OSINT) has become a significant role for various authorities and NGOs for advancing struggles in global health, human rights, and crisis management \citep{bernard_2018,evangelista_2021,kaufhold_2021}.
Following the view of OSINT as a tool, our work pursues the goal to support relief providers, government agencies, and other disaster-response stakeholders during ongoing and evolving crisis events.

We argue that NLP for disaster response can have a positive impact on comprehensive situational awareness and in decision-making processes such as coordination of particular services or physical goods. 
In the context of this work, positive impact means to supplement traditional information sources with social media streams that enable faster access to ongoing developments, first-party observations, and more fine-grained information content. 
For example, NLP for social media can enrich the information with the public as co-producers which may reveal critical subevents like missed or trapped people \citep{li_2018}.
Retrieving this kind of information could positively affect disaster management strategies and relief efforts during natural and human-made disasters.

In contrast, relying on social media as an information source runs the risk of introducing mis- and disinformation.
This can cause adverse effects on relief efforts and requires tailored strategies and particular care before the deployment of such models.
Furthermore, data privacy issues may arise due to the inherited properties of social media data.
Various anonymization processes should be taken into account for identifying and neutralizing sensitive references \citep{medlock_2006}.
In this work, the use of entity tokens as categorization can be seen as one kind of anonymization procedure.
However, model training with such entities could be task-specific and prone to error propagation by named entity recognition systems.

\section*{Acknowledgments}

The authors acknowledge the financial support by the Federal Ministry of Education and Research of Germany in the project ISAKI (project number 13N15572).
We also would like to thank the anonymous reviewers for their constructive feedback.

\bibliography{anthology,custom}

\begin{thebibliography}{42}
\expandafter\ifx\csname natexlab\endcsname\relax\def\natexlab#1{#1}\fi

\bibitem[{Alam et~al.(2018{\natexlab{a}})Alam, Joty, and Imran}]{alam_2018}
Firoj Alam, Shafiq Joty, and Muhammad Imran. 2018{\natexlab{a}}.
\newblock \href {https://doi.org/10.18653/v1/P18-1099} {Domain adaptation with
  adversarial training and graph embeddings}.
\newblock In \emph{Proceedings of the 56th Annual Meeting of the Association
  for Computational Linguistics (Volume 1: Long Papers)}, pages 1077--1087,
  Melbourne, Australia. Association for Computational Linguistics.

\bibitem[{Alam et~al.(2018{\natexlab{b}})Alam, Ofli, and
  Imran}]{crisismmd_2018}
Firoj Alam, Ferda Ofli, and Muhammad Imran. 2018{\natexlab{b}}.
\newblock Crisismmd: Multimodal twitter datasets from natural disasters.
\newblock In \emph{Proceedings of the 12th International AAAI Conference on Web
  and Social Media (ICWSM)}.

\bibitem[{Alam et~al.(2021)Alam, Sajjad, Imran, and Ofli}]{alam_2021}
Firoj Alam, Hassan Sajjad, Muhammad Imran, and Ferda Ofli. 2021.
\newblock Crisisbench: Benchmarking crisis-related social media datasets for
  humanitarian information processing.
\newblock In \emph{15th International Conference on Web and Social Media
  (ICWSM)}.

\bibitem[{Banerjee et~al.(2019)Banerjee, Akkaya, Perez-Sorrosal, and
  Tsioutsiouliklis}]{banerjee_2019}
Siddhartha Banerjee, Cem Akkaya, Francisco Perez-Sorrosal, and Kostas
  Tsioutsiouliklis. 2019.
\newblock \href {https://doi.org/10.18653/v1/P19-1633} {Hierarchical transfer
  learning for multi-label text classification}.
\newblock In \emph{Proceedings of the 57th Annual Meeting of the Association
  for Computational Linguistics}, pages 6295--6300, Florence, Italy.
  Association for Computational Linguistics.

\bibitem[{Ben-David et~al.(2020)Ben-David, Rabinovitz, and
  Reichart}]{ben_david_2020}
Eyal Ben-David, Carmel Rabinovitz, and Roi Reichart. 2020.
\newblock \href {https://doi.org/10.1162/tacl_a_00328} {{PERL}: Pivot-based
  domain adaptation for pre-trained deep contextualized embedding models}.
\newblock \emph{Transactions of the Association for Computational Linguistics},
  8:504--521.

\bibitem[{Bernard et~al.(2018)Bernard, Bowsher, Milner, Boyle, Patel, and
  Sullivan}]{bernard_2018}
Rose Bernard, G.~Bowsher, C.~Milner, P.~Boyle, P.~Patel, and R.~Sullivan. 2018.
\newblock \href {https://doi.org/10.1007/s10389-018-0899-3} {Intelligence and
  global health: assessing the role of open source and social media
  intelligence analysis in infectious disease outbreaks}.
\newblock \emph{Journal of Public Health}, 26(5):509--514.

\bibitem[{Brinkmann and Bizer(2021)}]{brinkmann_2021}
Alexander Brinkmann and Christian Bizer. 2021.
\newblock \href {https://madoc.bib.uni-mannheim.de/60102/} {Improving
  hierarchical product classification using domain-specific language
  modelling}.
\newblock \emph{Bulletin of the Technical Committee on Data Engineering / IEEE
  Computer Society}, 44(2):14--25.

\bibitem[{Buntain et~al.(2021)Buntain, McCreadie, and Soboroff}]{trec_is_2021}
Cody~L. Buntain, Richard McCreadie, and Ian Soboroff. 2021.
\newblock Incident {Streams} 2020: {TREC}-{IS} in the {Time} of {COVID}-19.
\newblock In \emph{ISCRAM 2021: 18th International Conference on Information
  Systems for Crisis Response and Management}.

\bibitem[{Clark et~al.(2020)Clark, Luong, Le, and Manning}]{electra_2020}
Kevin Clark, Minh-Thang Luong, Quoc~V. Le, and Christopher~D. Manning. 2020.
\newblock \href {https://openreview.net/pdf?id=r1xMH1BtvB} {{ELECTRA}:
  Pre-training text encoders as discriminators rather than generators}.
\newblock In \emph{ICLR}.

\bibitem[{Devlin et~al.(2019)Devlin, Chang, Lee, and Toutanova}]{devlin_2019}
Jacob Devlin, Ming-Wei Chang, Kenton Lee, and Kristina Toutanova. 2019.
\newblock \href {https://doi.org/10.18653/v1/N19-1423} {{BERT}: Pre-training of
  deep bidirectional transformers for language understanding}.
\newblock In \emph{Proceedings of the 2019 Conference of the North {A}merican
  Chapter of the Association for Computational Linguistics: Human Language
  Technologies, Volume 1 (Long and Short Papers)}, Minneapolis, Minnesota.
  Association for Computational Linguistics.

\bibitem[{Dumais and Chen(2000)}]{dumais_2000}
Susan Dumais and Hao Chen. 2000.
\newblock \href {https://doi.org/10.1145/345508.345593} {Hierarchical
  classification of web content}.
\newblock In \emph{Proceedings of the 23rd Annual International ACM SIGIR
  Conference on Research and Development in Information Retrieval}, SIGIR '00,
  page 256–263, New York, NY, USA. Association for Computing Machinery.

\bibitem[{Evangelista et~al.(2021)Evangelista, Sassi, Romero, and
  Napolitano}]{evangelista_2021}
João Rafael~Gonçalves Evangelista, Renato~José Sassi, Márcio Romero, and
  Domingos Napolitano. 2021.
\newblock \href {https://doi.org/10.1080/19361610.2020.1761737} {Systematic
  {Literature} {Review} to {Investigate} the {Application} of {Open} {Source}
  {Intelligence} ({OSINT}) with {Artificial} {Intelligence}}.
\newblock \emph{Journal of Applied Security Research}, 16(3):345--369.
\newblock Publisher: Routledge \_eprint:
  https://doi.org/10.1080/19361610.2020.1761737.

\bibitem[{Gururangan et~al.(2020)Gururangan, Marasovi{\'c}, Swayamdipta, Lo,
  Beltagy, Downey, and Smith}]{gururangan_2020}
Suchin Gururangan, Ana Marasovi{\'c}, Swabha Swayamdipta, Kyle Lo, Iz~Beltagy,
  Doug Downey, and Noah~A. Smith. 2020.
\newblock \href {https://doi.org/10.18653/v1/2020.acl-main.740} {Don{'}t stop
  pretraining: Adapt language models to domains and tasks}.
\newblock In \emph{Proceedings of the 58th Annual Meeting of the Association
  for Computational Linguistics}, pages 8342--8360, Online. Association for
  Computational Linguistics.

\bibitem[{Habdank et~al.(2017)Habdank, Rodehutskors, and Koch}]{habdank_2017}
Matthias Habdank, Nikolai Rodehutskors, and Rainer Koch. 2017.
\newblock \href {https://doi.org/10.1109/ICT-DM.2017.8275670} {Relevancy
  assessment of tweets using supervised learning techniques: Mining emergency
  related tweets for automated relevancy classification}.
\newblock In \emph{2017 4th International Conference on Information and
  Communication Technologies for Disaster Management (ICT-DM)}, pages 1--8.

\bibitem[{Howard and Ruder(2018)}]{howard_2018}
Jeremy Howard and Sebastian Ruder. 2018.
\newblock \href {https://doi.org/10.18653/v1/P18-1031} {Universal language
  model fine-tuning for text classification}.
\newblock In \emph{Proceedings of the 56th Annual Meeting of the Association
  for Computational Linguistics (Volume 1: Long Papers)}, pages 328--339,
  Melbourne, Australia. Association for Computational Linguistics.

\bibitem[{Kaufhold(2021)}]{kaufhold_2021}
Marc-André Kaufhold. 2021.
\newblock \href {https://doi.org/10.1007/978-3-658-33341-6} {\emph{Information
  {Refinement} {Technologies} for {Crisis} {Informatics}: {User} {Expectations}
  and {Design} {Principles} for {Social} {Media} and {Mobile} {Apps}}}.
\newblock Springer Fachmedien Wiesbaden, Wiesbaden.

\bibitem[{Kejriwal and Zhou(2019)}]{kejriwal_2019}
M.~Kejriwal and P.~Zhou. 2019.
\newblock \href {https://doi.org/10.1145/3341161.3342936} {Low-supervision
  urgency detection and transfer in short crisis messages}.
\newblock In \emph{2019 IEEE/ACM International Conference on Advances in Social
  Networks Analysis and Mining (ASONAM)}, pages 353--356, Los Alamitos, CA,
  USA. IEEE Computer Society.

\bibitem[{Kersten et~al.(2019)Kersten, Kruspe, Wiegmann, and
  Klan}]{kersten_2019}
Jens Kersten, Anna Kruspe, Matti Wiegmann, and Friederike Klan. 2019.
\newblock Robust filtering of crisis-related tweets.
\newblock In \emph{ISCRAM 2019: 16th International Conference on Information
  Systems for Crisis Response and Management}.

\bibitem[{Kruspe et~al.(2021)Kruspe, Kersten, and Klan}]{kruspe_2021}
Anna Kruspe, Jens Kersten, and Friederike Klan. 2021.
\newblock \href {https://doi.org/10.5194/nhess-21-1825-2021} {Review article:
  Detection of actionable tweets in crisis events}.
\newblock \emph{Natural Hazards and Earth System Sciences}, 21(6):1825--1845.

\bibitem[{Kumar et~al.(2011)Kumar, Barbier, Abbasi, and Liu}]{kumar_2011}
Shamanth Kumar, Geoffrey Barbier, Mohammad Abbasi, and Huan Liu. 2011.
\newblock \href {https://ojs.aaai.org/index.php/ICWSM/article/view/14079}
  {Tweettracker: An analysis tool for humanitarian and disaster relief}.
\newblock \emph{Proceedings of the International AAAI Conference on Web and
  Social Media}, 5(1):661--662.

\bibitem[{Li et~al.(2018)Li, Zhang, Tian, and Wang}]{li_2018}
Lifang Li, Qingpeng Zhang, Jun Tian, and Haolin Wang. 2018.
\newblock \href {https://doi.org/10.1016/j.ijinfomgt.2017.08.008}
  {Characterizing information propagation patterns in emergencies: {A} case
  study with {Yiliang} {Earthquake}}.
\newblock \emph{International Journal of Information Management}, 38(1):34--41.

\bibitem[{Liu et~al.(2019)Liu, Ott, Goyal, Du, Joshi, Chen, Levy, Lewis,
  Zettlemoyer, and Stoyanov}]{roberta_2019}
Yinhan Liu, Myle Ott, Naman Goyal, Jingfei Du, Mandar Joshi, Danqi Chen, Omer
  Levy, Mike Lewis, Luke Zettlemoyer, and Veselin Stoyanov. 2019.
\newblock \href {https://doi.org/10.48550/ARXIV.1907.11692} {Roberta: A
  robustly optimized bert pretraining approach}.

\bibitem[{McCreadie et~al.(2019)McCreadie, Buntain, and
  Soboroff}]{trec_is_2019}
Richard McCreadie, Cody~L. Buntain, and Ian Soboroff. 2019.
\newblock {TREC} {Incident} {Streams}: {Finding} {Actionable} {Information} on
  {Social} {Media}.
\newblock In \emph{ISCRAM 2019: 16th International Conference on Information
  Systems for Crisis Response and Management}.

\bibitem[{Medina~Maza et~al.(2020)Medina~Maza, Spiliopoulou, Hovy, and
  Hauptmann}]{medina_2020}
Salvador Medina~Maza, Evangelia Spiliopoulou, Eduard Hovy, and Alexander
  Hauptmann. 2020.
\newblock \href {https://doi.org/10.18653/v1/2020.findings-emnlp.344}
  {Event-related bias removal for real-time disaster events}.
\newblock In \emph{Findings of the Association for Computational Linguistics:
  EMNLP 2020}, pages 3858--3868, Online. Association for Computational
  Linguistics.

\bibitem[{Medlock(2006)}]{medlock_2006}
Ben Medlock. 2006.
\newblock \href {http://www.lrec-conf.org/proceedings/lrec2006/pdf/200_pdf.pdf}
  {An introduction to {NLP}-based textual anonymisation}.
\newblock In \emph{Proceedings of the Fifth International Conference on
  Language Resources and Evaluation ({LREC}{'}06)}, Genoa, Italy. European
  Language Resources Association (ELRA).

\bibitem[{Mehrotra et~al.(2022)Mehrotra, Mishra, and Pal}]{mehrotra_2022}
Harshit Mehrotra, Akanksha Mishra, and Sukomal Pal. 2022.
\newblock \href {https://doi.org/10.1007/s42979-021-00930-z} {A {Multi}-stage
  {Classification} {Framework} for {Disaster}-{Specific} {Tweets}}.
\newblock \emph{SN Computer Science}, 3(1):24.

\bibitem[{Mikolov et~al.(2013)Mikolov, Sutskever, Chen, Corrado, and
  Dean}]{mikolov_2013}
Tomas Mikolov, Ilya Sutskever, Kai Chen, Greg~S Corrado, and Jeff Dean. 2013.
\newblock \href
  {https://proceedings.neurips.cc/paper/2013/file/9aa42b31882ec039965f3c4923ce901b-Paper.pdf}
  {Distributed representations of words and phrases and their
  compositionality}.
\newblock In \emph{Advances in Neural Information Processing Systems},
  volume~26. Curran Associates, Inc.

\bibitem[{Miyazaki et~al.(2019)Miyazaki, Makino, Takei, Okamoto, and
  Goto}]{miyazaki_2019}
Taro Miyazaki, Kiminobu Makino, Yuka Takei, Hiroki Okamoto, and Jun Goto. 2019.
\newblock \href {https://doi.org/10.18653/v1/D19-1660} {Label embedding using
  hierarchical structure of labels for {T}witter classification}.
\newblock In \emph{Proceedings of the 2019 Conference on Empirical Methods in
  Natural Language Processing and the 9th International Joint Conference on
  Natural Language Processing (EMNLP-IJCNLP)}, pages 6317--6322, Hong Kong,
  China. Association for Computational Linguistics.

\bibitem[{Nguyen et~al.(2020)Nguyen, Vu, and Tuan~Nguyen}]{bertweet_2020}
Dat~Quoc Nguyen, Thanh Vu, and Anh Tuan~Nguyen. 2020.
\newblock \href {https://doi.org/10.18653/v1/2020.emnlp-demos.2} {{BERT}weet: A
  pre-trained language model for {E}nglish tweets}.
\newblock In \emph{Proceedings of the 2020 Conference on Empirical Methods in
  Natural Language Processing: System Demonstrations}, pages 9--14, Online.
  Association for Computational Linguistics.

\bibitem[{Poblet et~al.(2014)Poblet, Garc{\'i}a-Cuesta, and
  Casanovas}]{poblet_2014}
Marta Poblet, Esteban Garc{\'i}a-Cuesta, and Pompeu Casanovas. 2014.
\newblock Crowdsourcing tools for disaster management: A review of platforms
  and methods.
\newblock In \emph{AI Approaches to the Complexity of Legal Systems}, pages
  261--274, Berlin, Heidelberg. Springer Berlin Heidelberg.

\bibitem[{Reuter et~al.(2018)Reuter, Hughes, and Kaufhold}]{reuter_2018}
Christian Reuter, Amanda~Lee Hughes, and Marc-André Kaufhold. 2018.
\newblock \href {https://doi.org/10.1080/10447318.2018.1427832} {Social {Media}
  in {Crisis} {Management}: {An} {Evaluation} and {Analysis} of {Crisis}
  {Informatics} {Research}}.
\newblock \emph{International Journal of Human–Computer Interaction},
  34(4):280--294.

\bibitem[{Sakaki et~al.(2010)Sakaki, Okazaki, and Matsuo}]{sakaki_2010}
Takeshi Sakaki, Makoto Okazaki, and Yutaka Matsuo. 2010.
\newblock \href {https://doi.org/10.1145/1772690.1772777} {Earthquake shakes
  twitter users: Real-time event detection by social sensors}.
\newblock In \emph{Proceedings of the 19th International Conference on World
  Wide Web}, WWW '10, page 851–860, New York, NY, USA. Association for
  Computing Machinery.

\bibitem[{Sarmiento and Poblete(2021)}]{sarmiento_2021}
Hernan Sarmiento and Barbara Poblete. 2021.
\newblock \href {https://doi.org/10.1145/3412841.3442044} {Crisis
  communication: A comparative study of communication patterns across crisis
  events in social media}.
\newblock In \emph{Proceedings of the 36th Annual ACM Symposium on Applied
  Computing}, SAC '21, page 1711–1720, New York, NY, USA. Association for
  Computing Machinery.

\bibitem[{Segura-Bedmar et~al.(2013)Segura-Bedmar, Martínez, and
  Herrero-Zazo}]{semeval_2013}
Isabel Segura-Bedmar, Paloma Martínez, and María Herrero-Zazo. 2013.
\newblock \href {https://aclanthology.org/S13-2056} {{SemEval}-2013 {Task} 9 :
  {Extraction} of {Drug}-{Drug} {Interactions} from {Biomedical} {Texts}
  ({DDIExtraction} 2013)}.
\newblock In \emph{Second {Joint} {Conference} on {Lexical} and {Computational}
  {Semantics} (*{SEM}), {Volume} 2: {Proceedings} of the {Seventh}
  {International} {Workshop} on {Semantic} {Evaluation} ({SemEval} 2013)},
  pages 341--350, Atlanta, Georgia, USA. Association for Computational
  Linguistics.

\bibitem[{Snyder et~al.(2019)Snyder, Lin, Karimzadeh, Goldwasser, and
  Ebert}]{snyder_2019}
Luke~S. Snyder, Yi-Shan Lin, Morteza Karimzadeh, Dan Goldwasser, and David~S.
  Ebert. 2019.
\newblock \href {https://doi.org/10.1109/TVCG.2019.2934614} {Interactive
  {Learning} for {Identifying} {Relevant} {Tweets} to {Support} {Real}-time
  {Situational} {Awareness}}.
\newblock \emph{IEEE Transactions on Visualization and Computer Graphics},
  pages 1--1.

\bibitem[{Vaswani et~al.(2017)Vaswani, Shazeer, Parmar, Uszkoreit, Jones,
  Gomez, Kaiser, and Polosukhin}]{vaswani_2017}
Ashish Vaswani, Noam Shazeer, Niki Parmar, Jakob Uszkoreit, Llion Jones,
  Aidan~N Gomez, \L~ukasz Kaiser, and Illia Polosukhin. 2017.
\newblock \href
  {https://proceedings.neurips.cc/paper/2017/file/3f5ee243547dee91fbd053c1c4a845aa-Paper.pdf}
  {Attention is all you need}.
\newblock In \emph{Advances in Neural Information Processing Systems},
  volume~30. Curran Associates, Inc.

\bibitem[{Wang et~al.(2021)Wang, Nulty, and Lillis}]{wang_2021a}
Congcong Wang, Paul Nulty, and David Lillis. 2021.
\newblock {Transformer-based Multi-task Learning for Disaster Tweet
  Categorisation}.
\newblock In \emph{ISCRAM 2021: 18th International Conference on Information
  Systems for Crisis Response and Management}.

\bibitem[{Wehrmann et~al.(2018)Wehrmann, Cerri, and Barros}]{wehrmann_2018}
Jonatas Wehrmann, Ricardo Cerri, and Rodrigo Barros. 2018.
\newblock \href {https://proceedings.mlr.press/v80/wehrmann18a.html}
  {Hierarchical multi-label classification networks}.
\newblock In \emph{Proceedings of the 35th International Conference on Machine
  Learning}, volume~80 of \emph{Proceedings of Machine Learning Research},
  pages 5075--5084. PMLR.

\bibitem[{Wiegmann et~al.(2020)Wiegmann, Kersten, Klan, Potthast, and
  Stein}]{wiegmann_2020}
Matti Wiegmann, Jens Kersten, Friederike Klan, Martin Potthast, and Benno
  Stein. 2020.
\newblock \href {https://doi.org/10.5281/ZENODO.3713920} {Analysis of
  {Detection} {Models} for {Disaster}-{Related} {Tweets}}.
\newblock In \emph{ISCRAM 2020: 17th International Conference on Information
  Systems for Crisis Response and Management}.

\bibitem[{Wolf et~al.(2020)Wolf, Debut, Sanh, Chaumond, Delangue, Moi, Cistac,
  Rault, Louf, Funtowicz, Davison, Shleifer, von Platen, Ma, Jernite, Plu, Xu,
  Le~Scao, Gugger, Drame, Lhoest, and Rush}]{wolf_2020}
Thomas Wolf, Lysandre Debut, Victor Sanh, Julien Chaumond, Clement Delangue,
  Anthony Moi, Pierric Cistac, Tim Rault, Remi Louf, Morgan Funtowicz, Joe
  Davison, Sam Shleifer, Patrick von Platen, Clara Ma, Yacine Jernite, Julien
  Plu, Canwen Xu, Teven Le~Scao, Sylvain Gugger, Mariama Drame, Quentin Lhoest,
  and Alexander Rush. 2020.
\newblock \href {https://doi.org/10.18653/v1/2020.emnlp-demos.6} {Transformers:
  State-of-the-art natural language processing}.
\newblock In \emph{Proceedings of the 2020 Conference on Empirical Methods in
  Natural Language Processing: System Demonstrations}, pages 38--45, Online.
  Association for Computational Linguistics.

\bibitem[{Yu et~al.(2021)Yu, Sun, Sun, and Liu}]{yu_2021}
Yipeng Yu, Zixun Sun, Chi Sun, and Wenqiang Liu. 2021.
\newblock \href {https://doi.org/10.1109/ICTAI52525.2021.00180} {Hierarchical
  multilabel text classification via multitask learning}.
\newblock In \emph{2021 IEEE 33rd International Conference on Tools with
  Artificial Intelligence (ICTAI)}, pages 1138--1143.

\bibitem[{Zahera et~al.(2021)Zahera, Jalota, Sherif, and Ngomo}]{zahera_2021}
Hamada~M. Zahera, Rricha Jalota, Mohamed~Ahmed Sherif, and Axel-Cyrille~Ngonga
  Ngomo. 2021.
\newblock \href {https://doi.org/10.1109/ACCESS.2021.3107812} {I-aid:
  Identifying actionable information from disaster-related tweets}.
\newblock \emph{IEEE Access}, 9:118861--118870.

\end{thebibliography}
\bibliographystyle{acl_natbib}

\appendix

\section{Overview of Information Types}
\label{appendix:dataset_details}

We list all information types of the TREC-IS dataset in \tablename~\ref{table:appendices:dataset}.
The value in the last column indicates the number of Twitter posts to which the corresponding labels were assigned.
\tablename~\ref{table:appendices:examples} displays example tweets for various events with the corresponding labels from the TREC-IS dataset.

\section{Hyper-Parameters}
\label{appendix:hyperparameters}

The search space for TF-IDF+LR included ngram-range, max features and regularization strength.
In terms of BERT fine-tuning, we manually experimented with the same parameters as in \citet{wang_2021a} and selected in line with this work the learning rate $5e-5$ and batch size $32$. 
Due to computational constraints, we used for BERT pre-training the TAPT parameters of \citet{gururangan_2020}. Similar to \citet{ben_david_2020}, we experimented with the MLM probabilities $\alpha \in \{0.1,0.3,0.5,0.8\}$ and $\beta \in \{0.1,0.3,0.5,0.8\}$ and found the setup $\alpha=0.5$ and $\beta=0.1$ to perform best. 
This is in line with \citet{ben_david_2020} which empirically show good results. For MTL we tuned $\lambda \in \{0.1, 0.5, 0.9\}$ and finally set $\lambda=0.1$. 
We trained all transformer models with the \textit{Transformers} library \citep{wolf_2020} and \textit{AdamW} for up to 50 (pre-training) and 15 (fine-tuning) epochs, evaluated the performance each 1000 steps on the development set and selected the best performing checkpoint. 
If not other mentioned, we used for the rest of the hyper-parameters the default setup of BERT$_{BASE}$ from the \textit{Transformers} library.

\begin{table*}[t]
    \centering
    \begin{tabular}{lllll}
        \hline
        \textbf{Id} & \textbf{Upper Class (L$_{T}$)}  & \textbf{Lower Class (L$_{B}$)} & \textbf{Actionable (AIT)} & \textbf{\# tweets}\\
        \hline
        RQ 01 & Request & GoodsServices & \multicolumn{1}{c}{\checkmark} & 194 \\
        RQ 02 & Request & InformationWanted & & 395 \\
        RQ 03 & Request & SearchAndRescue & \multicolumn{1}{c}{\checkmark} & 274 \\
        CTA 01 & CallToAction & Donations & & 986 \\
        CTA 02 & CallToAction & MovePeople & \multicolumn{1}{c}{\checkmark} & 679 \\
        CTA 03 & CallToAction & Volunteer & & 242 \\
        O 01 & Other & Advice & & 3,277 \\
        O 02 & Other & ContextualInformation & & 4,583 \\
        O 03 & Other & Discussion & & 5,303 \\
        O 04 & Other & Irrelevant & & 23,053 \\
        O 05 & Other & Sentiment & & 11,101 \\
        RP 01 & Report & CleanUp & & 493 \\
        RP 02 & Report & EmergingThreats & \multicolumn{1}{c}{\checkmark} & 6,930 \\
        RP 03 & Report & Factoid & & 10,224 \\
        RP 04 & Report & NewSubEvent & \multicolumn{1}{c}{\checkmark} & 2,806 \\
        RP 05 & Report & FirstPartyObservation & & 5,290 \\
        RP 06 & Report & Hashtags & & 15,787 \\
        RP 07 & Report & Location & & 23,676 \\
        RP 08 & Report & MultimediaShare & & 22,976 \\
        RP 09 & Report & News & & 18,374 \\
        RP 10 & Report & Official & & 2,836 \\
        RP 11 & Report & OriginalEvent & & 4,148 \\
        RP 12 & Report & ServiceAvailable & \multicolumn{1}{c}{\checkmark} & 2,184 \\
        RP 13 & Report & ThirdPartyObservation & & 17,223 \\
        RP 14 & Report & Weather & & 7,655\\
        \hline
    \end{tabular}
    \caption{Information types and hierarchical structure of labels.}
    \label{table:appendices:dataset}
\end{table*}

\begin{table*}[t]
    \centering
    \rowcolors{2}{white}{gray!10}
    \begin{tabular}{lp{0.25\linewidth}p{0.4\linewidth}}
        \hline
        \textbf{Event} & \textbf{Labels}  & \textbf{Tweet} \\
        \hline
        Wildfire Colorado 2012 & Irrelevant & From the train, showing the smoke filled sky from the \#Lithgow \#nswfires \\
        Bushfire Australia 2013 & ThirdPartyObservation, Factoid, Advice & FIRE UPDATE: Families told to be ready to run as a massive 300km wall of fire sweeps through Blue Mtns. \#nswfires \\
        Earthquake Chile 2014 & News & New this morning: At least 6 people are dead after the massive M8.2 quake in \#Chile \\
        Explosion Beirut 2020 & Location, Factoid, OriginalEvent, ContextualInformation & At least 25 dead and more than 2,500 injured as a result of the Beirut Port explosion according to the Lebanese Health Ministry \\
        Flood Colorado 2013 & Factoid & 5 people confirmed dead in Colorado flooding, and 1,254 people unaccounted for statewide, official says \\
        Hurricane Florence 2018 & Weather, Location, Hashtags & We have 2.5 inches here 2.6 miles north-west of Downtown awake Forest. \#FlorenceHurricane2018 \\
        \hline
    \end{tabular}
    \caption{Example tweets and labels for different events.}
    \label{table:appendices:examples}
\end{table*}

\end{document}